\documentclass[sigconf,nonacm]{acmart}
\usepackage{multirow}

\newcommand{\TopR}{\operatorname{TopR}}
\newcommand{\TopK}{\operatorname{TopK}}
\newcommand{\softmax}{\operatorname{softmax}}

\AtBeginDocument{%
  }

\setcopyright{none}
\begin{document}

\title[Consistency and Conflict across MLLM Layers]{Decoupling Consistency and Conflict across MLLM Layers for Fake News Detection}

\author{Kai Qian}
\affiliation{
  \department{School of Computer Science and Technology}
  \institution{Soochow University}
  \country{China}}
\authornote{These authors contributed equally to this work.}
\author{Dongjie Xu}
\affiliation{
  \institution{Soochow University}
  \country{China}}
\authornotemark[1]
\author{Weijie Shi}
\affiliation{
  \institution{Hong Kong University of Science and Technology}
  \country{China}}
\author{Jiaqi Wang}
\affiliation{
  \department{Computer Science and Engineering}
  \institution{The Chinese University of Hong Kong}
  \country{China}}
\author{Haolun Ma}
\affiliation{
  \department{DSA}
  \institution{The Hong Kong University of Science and Technology}
  \country{China}}
\author{Jia Zhu}
\affiliation{
  \department{School of Education}
  \institution{Zhejiang Normal University}
  \country{China}}
\author{Jiajie Xu}
\affiliation{
  \department{School of Computer Science and Technology}
  \institution{Soochow University}
  \country{China}}
\renewcommand{\shortauthors}{Qian et al.}

\begin{abstract}
The trustworthiness of Web content faces a growing threat from multimodal fake news. Real news usually shows aligned image-text semantics, while fake news often contains a few conflicts at varying levels, such as mismatches in objects, scenes, events, and claims. However, existing methods usually treat consistency and conflict in a coarse and static way, limiting their ability to capture sparse conflicts and adapt to varying conflict levels. Inspired by the distinct knowledge encoded across layers in large language models, we propose Multi-Level Consistency and Conflict Decoupling (MLCCD), which captures conflicts at different levels from different layers of multimodal large language models (MLLMs). Specifically, we first use the MLLM's final decision representation to select visual and textual tokens relevant to the task, preventing irrelevant content from diluting sparse conflicts. Then, we preserve tokens whose representations change sharply between adjacent layers, capturing the knowledge each layer adds. Finally, we decouple this knowledge into consistency and conflict evidence and aggregate each across layers with its own router, adapting to varying conflict levels. Experiments on three benchmarks show that MLCCD outperforms existing methods.
\end{abstract}



\maketitle

\section{Introduction}
The Web has become a major medium for accessing and sharing news. However, the rapid dissemination of fake news through online platforms poses a serious threat to the credibility of Web content\cite{khattar2019mvae}.%
\begingroup
\renewcommand{\thefootnote}{$\dagger$}%
\footnote{\textit{Relevance to the Web:} Fake news on Web platforms often pairs authentic images with misleading text, so the decisive evidence lies in a few image-text conflicts within otherwise consistent posts. MLCCD mines such conflicts from multimodal Web posts in Chinese and English using the internal layers of MLLMs. This aligns with the Web Mining, Multimedia and Multilingual Content Analysis track's focus on mining and understanding multimedia Web content with large pretrained models.}%
\addtocounter{footnote}{-1}%
\endgroup As news on these platforms commonly combines text and images \cite{cheema2023understanding}, multimodal fake news detection is an important part of addressing this threat.

In multimodal news, misleading information can arise from how images and text are combined, even when each modality appears plausible on its own \cite{cheema2023understanding,qi2024sniffer}. Semantic consistency between the two modalities therefore provides useful evidence for verification \cite{xue2021detecting,muller2020multimodal}, but broad alignment can coexist with conflicts in specific details. For example, the text may misidentify a person or object shown in the image \cite{qi2021improving}, or describe a setting that differs from the depicted scene \cite{muller2020multimodal}. Such mismatches can also concern the reported event, with the image and text referring to different events despite sharing the same entities \cite{ma2024event}. Beyond the depicted content, an authentic image may be used to support a claim that contradicts its original context \cite{qi2024sniffer}. Together, these examples illustrate how broad semantic agreement can coexist with decisive conflicts involving only a few details at different semantic levels.

\begin{figure}[t]
  \centering
  \includegraphics[width=\columnwidth]{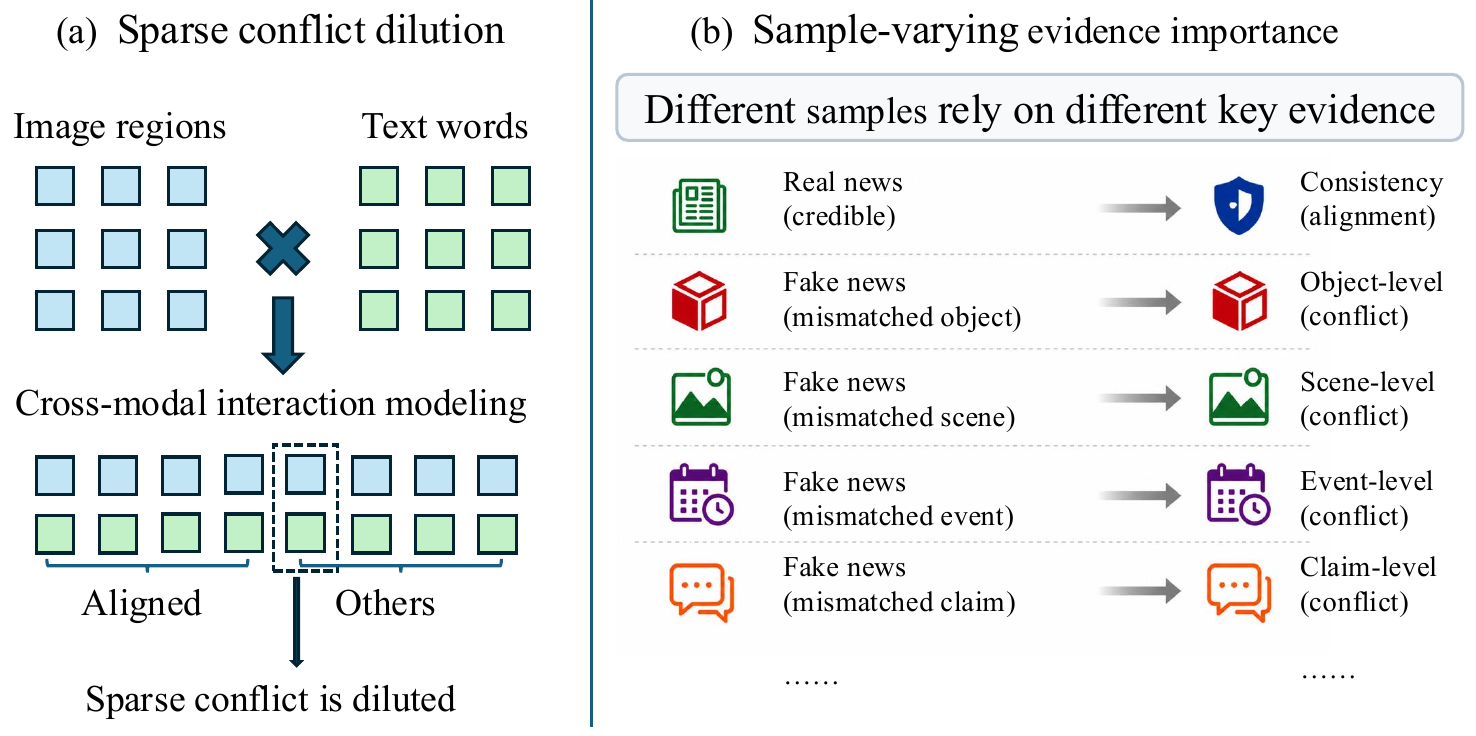}
  \caption{
  Illustration of two challenges in multimodal fake news detection.
  (a) Sparse conflicts can be diluted when aggregated with other content.
  (b) The importance of consistency and conflict evidence can vary across samples and semantic levels.
  }
  \label{fig:motivation}
\end{figure}

To capture these semantic relations, existing studies have explored a range of modeling strategies for multimodal fake news detection, including contrastive alignment \cite{shen2024multimodal}, semantic consistency modeling \cite{tao2025multimodal}, and higher order interaction learning \cite{sun2025multimodal}. Nevertheless, those methods still model consistency and conflict at a relatively coarse granularity, without explicitly adapting evidence aggregation across semantic levels to individual samples. These limitations highlight the following two key challenges.

\textbf{First, sparse conflict clues are easily diluted.}
As illustrated in Figure~\ref{fig:motivation}(a), a conflict may involve only a few interactions between image regions and text, while most other interactions reflect semantic agreement or involve content unrelated to verification. If these interactions are aggregated without sufficiently distinguishing their value for verification, the more abundant signals can dominate the resulting representation and obscure the few conflicting details. Therefore, it is necessary to preserve the details that reveal conflicts while reducing interference from content unrelated to verification.

\textbf{Second, the importance of consistency and conflict evidence varies across samples.}
As illustrated in Figure~\ref{fig:motivation}(b), consistency between images and text can provide supporting evidence for real news, while the decisive evidence for fake news may involve conflicts concerning objects, scenes, events, or claims. Assigning fixed importance to different forms of evidence may fail to accommodate these variations, giving insufficient emphasis to the clues most informative for a particular sample. Therefore, it is necessary to adapt evidence fusion to each sample so that the most informative consistency and conflict clues receive sufficient emphasis.

To address these challenges, we turn to representations from different layers of multimodal large language models (MLLMs). Prior studies of large language models (LLMs) have identified specific layers involved in factual recall \cite{meng2022locating} and shown that contrasting outputs from different layers can help reveal factual knowledge \cite{chuang2024dola}. Motivated by these findings, we propose Multi-Level Consistency and Conflict Decoupling (MLCCD) to extract consistency and conflict evidence from different MLLM layers and adapt its aggregation to the needs of individual samples. Specifically, we first use the MLLM's final decision representation as a query to select visual and textual tokens relevant to the task, aiming to reduce interference from irrelevant content. Among these tokens, we then retain those whose representations change sharply between adjacent layers to capture the information introduced at each layer. Finally, we decouple the retained representations at each layer into consistency and conflict evidence and aggregate each type across layers with its own router to adapt to varying evidence needs across samples.

Our main contributions are summarized as follows:
\begin{itemize}
    \item We present a perspective for multimodal fake news detection that draws on knowledge from different MLLM layers to capture consistency and conflict at multiple levels.

    \item We propose MLCCD, a Multi-Level Consistency and Conflict Decoupling framework. It extracts information relevant to verification to preserve sparse conflict clues, and decouples and adaptively aggregates consistency and conflict evidence to accommodate varying evidence needs across samples.

    \item Experiments on three benchmarks show that MLCCD outperforms the compared baselines.
\end{itemize}

\section{Related Work}

\subsection{Multimodal Fake News Detection}

Multimodal fake news detection aims to verify news credibility by jointly considering textual content, visual information, and their semantic relations \cite{giachanou2020multimodal}. Since mismatches between text and images can provide clues to fake news \cite{xue2021detecting}, existing studies have explored approaches such as joint multimodal representation learning \cite{khattar2019mvae}, cross modal alignment \cite{wang2023cross}, consistency modeling \cite{xue2021detecting}, modality interaction modeling \cite{liu2025modality}, and knowledge enhanced verification \cite{gao2024knowledge}. Recent methods further employ multimodal large language models to support explanation generation \cite{qi2024sniffer} and reasoning with generated evidence \cite{yan2025debunk}. However, these methods rarely exploit the knowledge encoded across MLLM layers and usually model consistency and conflict in a coarse or static way, making it difficult to capture sparse conflicts or adapt to varying conflict levels.

\subsection{Multimodal Large Language Models}

Multimodal large language models (MLLMs) extend large language models (LLMs) to multimodal perception and reasoning by connecting visual encoders with language models through modality alignment modules \cite{bai2025qwen3,grattafiori2024llama}. With large scale vision language pretraining \cite{chen2023vlp} and multimodal instruction tuning \cite{zhao2023svit}, MLLMs support visual question answering \cite{liu2024improved}, image captioning \cite{sarto2025image}, visual grounding \cite{xiao2025towards}, instruction following \cite{huang2025visual}, and complex multimodal understanding \cite{yin2024survey}. Beyond their final outputs, intermediate layers also carry useful information. Recent studies show that different layers of the visual encoder contribute differently to multimodal understanding, and that multilayer visual feature fusion can improve generalization \cite{lin2025multi,li2026instruction,chen2025rethinking}. In LLMs, intermediate representations can provide useful features for downstream tasks \cite{skean2025layer}, specific layers are involved in factual recall \cite{meng2022locating}, and contrasting outputs across layers can help reveal factual knowledge \cite{chuang2024dola}. However, the visual encoder sees only the image, while image and text tokens first interact in the language backbone. We therefore investigate consistency and conflict evidence across layers of the MLLM language backbone.

\section{Preliminaries}

\begin{figure*}[t]
  \centering
  \includegraphics[width=\textwidth]{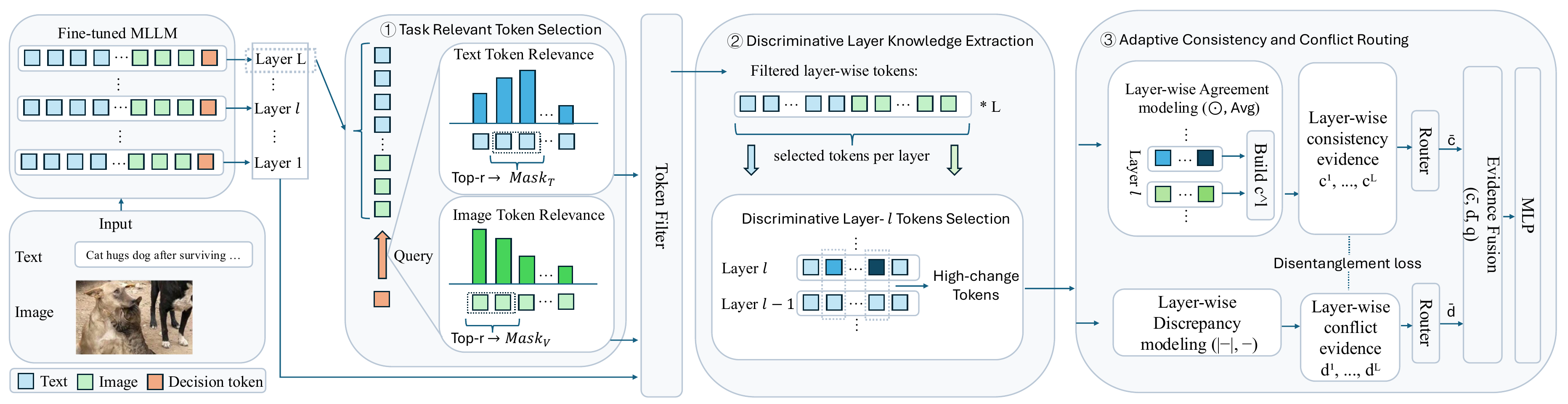}
  \caption{Overview of MLCCD. The frozen LoRA tuned MLLM provides layer wise hidden states, and the proposed evidence network selects task relevant tokens, extracts discriminative layer knowledge, and adaptively routes consistency and conflict evidence for fake news prediction.}
  \label{fig:framework}
\end{figure*}

\subsection{Task Definition}

For the \(i\)-th multimodal news post, let \(X_i^{t}\) denote its textual content, \(X_i^{v}\) denote its visual content, and \(y_i\in\{0,1\}\) denote its ground truth label, where \(y_i=1\) indicates fake news and \(y_i=0\) indicates real news. Given \(N\) training examples, the dataset is denoted as \(\mathcal{D}=\{(X_i^{t},X_i^{v},y_i)\}_{i=1}^{N}\). The goal of multimodal fake news detection is to learn a prediction function \(f_{\theta}\) that estimates the fake news probability \(\hat{y}_i=f_{\theta}(X_i^{t},X_i^{v})\) for each post.

\subsection{MLLM Layer Representations}

We use a LoRA tuned MLLM as the feature backbone. After tuning, the backbone is frozen, and the proposed evidence network operates on its hidden states. For each input post, the backbone produces layer wise token representations \(\mathcal{H}=\{H^1,H^2,\ldots,H^L\}\). For layer \(\ell\), the hidden sequence \(H^\ell=\{h_i^{\ell}\}_{i=1}^{T}\) consists of \(T\) token representations, where \(L\) is the number of MLLM layers and \(h_i^{\ell}\) is the hidden state of the \(i\)-th token in the input multimodal sequence at layer \(\ell\). We use two binary modality masks, \(M^t\in\{0,1\}^{T}\) and \(M^v\in\{0,1\}^{T}\), to distinguish textual and visual tokens. Specifically, \(M_i^t=1\) indicates that position \(i\) belongs to the text modality, and \(M_i^v=1\) indicates that position \(i\) belongs to the image modality. Viewing binary masks as sets of selected positions, the valid token mask is defined as their union, \(M=M^t\lor M^v\).

\section{Methodology}

\subsection{Overview}

MLCCD detects multimodal fake news by transforming the internal layer representations of an MLLM into structured verification evidence. Given a news post with an image and text, the MLLM encodes both modalities into a unified hidden sequence, where textual and visual elements are treated as tokens. Since decisive fake news clues are often sparse and may appear at different semantic depths, MLCCD progressively constructs evidence from token selection to layer knowledge extraction, and then performs fine-grained consistency and conflict decoupling across MLLM layers.

As shown in Figure~\ref{fig:framework}, MLCCD consists of three modules. \textit{Task Relevant Token Selection} uses the final decision representation as a query to select visual and textual tokens related to the verification decision. \textit{Discriminative Layer Knowledge Extraction} retains selected tokens with distinctive representation changes between adjacent layers, capturing layer specific knowledge from the MLLM hierarchy. \textit{Adaptive Consistency and Conflict Routing} builds separate evidence streams for cross modal agreement and discrepancy, and adaptively aggregates them across layers with independent routers.

\subsection{Task Relevant Token Selection}

This module constructs a compact evidence pool before layer knowledge extraction. Multimodal fake news clues are often localized in a small subset of textual or visual tokens, while many other tokens provide background information that is related to the post but less useful for verification. Directly aggregating all tokens may therefore dilute the decisive clues. To address this issue, we use the MLLM's decision context to measure token relevance and select informative textual and visual tokens separately.

In our prompt format, the MLLM is instructed to output the verification label as the first generated answer token. For an autoregressive MLLM, this answer is predicted from the hidden state of the final valid prompt token. Let \(i^\star\) denote the position of this token and \(q_0=h_{i^\star}^{L}\) denote its hidden state in the last MLLM layer. This hidden state provides a compact decision context for the verification task, and we use it as a task query. To identify tokens that are most related to the pending verification answer, we compare this query with each final layer token in a shared evidence space:
\begin{equation}
s_i=
\frac{\phi_h(h_i^L)^\top \phi_q(q_0)}
{\sqrt{d}}.
\label{eq:decision-relevance}
\end{equation}
Here, \(\phi_h\) and \(\phi_q\) are learnable projection functions, each consisting of layer normalization followed by a linear layer, and \(d\) is their output dimension. For brevity, we denote the projected query \(\phi_q(q_0)\) as \(q\) in subsequent modules. A larger \(s_i\) indicates that token \(i\) is more relevant to the verification decision and should be more likely to enter the evidence pool.

We then perform modality wise selection based on the relevance scores. For visual tokens, the candidate mask is \(C^v=M^v\). For textual tokens, the decision token is removed from the candidate set, so \(C^t=M^t\setminus\{i^\star\}\). For each modality \(m\in\{t,v\}\), the selected token mask \(\tilde{M}^{m}\) is obtained by top ratio selection:
\begin{equation}
\tilde{M}^{m}
=
\TopR(s,C^m,\rho_q^m),
\quad
m\in\{t,v\},
\label{eq:task-token-selection}
\end{equation}
where \(\TopR(s,C^m,\rho_q^m)\) keeps the highest scored \(\rho_q^m\) proportion of tokens within the candidate mask \(C^m\). The keep ratio \(\rho_q^m\) is modality dependent because the numbers of textual and visual tokens can differ substantially. This design allows each modality to preserve an appropriate amount of task relevant information and prevents the modality with more tokens from dominating the evidence pool. We also compute the normalized relevance distribution \(\pi_i^m\) over selected tokens:
\begin{equation}
\pi_i^{m}
=
\softmax_{i\in\tilde{M}^{m}}(s_i).
\label{eq:selection-relevance}
\end{equation}
The relevance distribution \(\pi_i^m\) is later used as a reliability cue for layer wise evidence extraction.

\subsection{Discriminative Layer Knowledge Extraction}

After task relevant tokens are selected, this module further identifies which selected tokens carry discriminative layer knowledge. Different MLLM layers provide different representations of the input, and a token whose representation changes strongly between adjacent layers is likely to contain layer specific evidence. Therefore, we use representation changes across layers to preserve tokens that are both task relevant and layer distinctive.

For token \(i\) at layer \(\ell\), the layer change score \(e_i^\ell\) measures how much its representation changes from the previous layer. Since the first layer has no preceding layer, its score is set to zero as a boundary convention:
\begin{equation}
e_i^{\ell}
=
\begin{cases}
0, & \ell=1,\\
\|h_i^{\ell}-h_i^{\ell-1}\|_2, & \ell>1.
\end{cases}
\label{eq:layer-change}
\end{equation}
For each modality \(m\) and each layer \(\ell\), we standardize the layer change scores within the selected candidate tokens \(\tilde{M}^{m}\) to obtain \(e_{i,\mathrm{std}}^{m,\ell}\), which measures the relative distinctiveness of token \(i\) within modality \(m\) at layer \(\ell\). The discriminative token mask \(\bar{M}^{m,\ell}\) keeps the selected tokens with the largest standardized changes:
\begin{equation}
\bar{M}^{m,\ell}
=
\TopR(e_{\mathrm{std}}^{m,\ell},\tilde{M}^{m},\rho_e),
\quad
m\in\{t,v\}.
\label{eq:layer-token-selection}
\end{equation}
Here, \(\rho_e\) is the layer evidence keep ratio.

To integrate task relevance and layer distinctiveness, each retained token receives a weight \(a_i^{m,\ell}\):
\begin{equation}
a_i^{m,\ell}
=
\softmax_{i\in\bar{M}^{m,\ell}}(s_i+\gamma e_{i,\mathrm{std}}^{m,\ell}),
\label{eq:token-weight}
\end{equation}
The scalar \(\gamma\) is a learnable positive scale. The modality representation \(u_m^\ell\) at layer \(\ell\) is obtained by weighted pooling:
\begin{equation}
u_m^{\ell}
=
\sum_{i\in\bar{M}^{m,\ell}}a_i^{m,\ell}h_i^{\ell},
\quad
m\in\{t,v\}.
\label{eq:modality-pooling}
\end{equation}
To preserve modality identity and layer order, we project \(u_m^\ell\) with a learnable projection function \(\phi_f\), and combine the result with the layer position embedding \(p^\ell\) and the modality embedding \(b_m\):
\begin{equation}
\hat{z}_m^{\ell}
=
\phi_f(u_m^{\ell})+p^{\ell}+b_m.
\label{eq:layer-input}
\end{equation}
Here, \(\phi_f\) consists of layer normalization, a linear layer, GELU, and dropout.
The resulting \(\hat{z}_t^\ell\) and \(\hat{z}_v^\ell\) are layer aware textual and visual representations that preserve task relevance and discriminative layer knowledge.
In addition, the token weights are also used to summarize the reliability of the extracted evidence. Specifically, \(\mu_m^\ell\) measures retained decision relevance, and \(\nu_m^\ell\) measures retained layer knowledge strength:
\begin{equation}
\begin{aligned}
\mu_m^{\ell}
&=
\sum_{i\in\bar{M}^{m,\ell}}a_i^{m,\ell}\pi_i^m,
\\
\nu_m^{\ell}
&=
\sum_{i\in\bar{M}^{m,\ell}}a_i^{m,\ell}\sigma(e_{i,\mathrm{std}}^{m,\ell}).
\end{aligned}
\label{eq:modality-reliability}
\end{equation}
These reliability scores are passed to the routing module together with the layer wise representations.

\subsection{Adaptive Consistency and Conflict Routing}

Given the extracted layer wise representations, MLCCD first applies a lightweight refinement before constructing consistency and conflict evidence. Specifically, a Transformer encoder \(E_{\mathrm{layer}}\) is applied along the layer dimension to slightly contextualize each layer representation while keeping the layer wise outputs separate:
\begin{equation}
z_m^{1:L}
=
E_{\mathrm{layer}}(\hat{z}_m^{1:L}).
\label{eq:layer-encoder}
\end{equation}
Based on the refined representations \(z_t^\ell\) and \(z_v^\ell\), MLCCD separates cross modal agreement from cross modal discrepancy at each layer. Consistency evidence reflects whether the visual and textual modalities support compatible semantics, while conflict evidence captures their divergent signals. The two evidence streams are then routed independently according to their layer wise reliability. In this way, decoupling is performed at multiple semantic levels instead of only on a single global multimodal representation.

For each layer \(\ell\), the consistency stream represents cross modal agreement with element wise interaction and averaged context:
\begin{equation}
g_c^{\ell}
=
\left[
z_t^{\ell}\odot z_v^{\ell};
\frac{z_t^{\ell}+z_v^{\ell}}{2}
\right].
\label{eq:consistency-input}
\end{equation}
The conflict stream represents cross modal discrepancy with absolute and directional differences:
\begin{equation}
g_d^{\ell}
=
\left[
|z_t^{\ell}-z_v^{\ell}|;
z_t^{\ell}-z_v^{\ell}
\right].
\label{eq:conflict-input}
\end{equation}
The branch inputs from all layers are then transformed by feed forward builders \(\phi_c\) and \(\phi_d\), followed by branch encoders \(E_c\) and \(E_d\), producing layer wise consistency evidence \(c^{1:L}\) and conflict evidence \(d^{1:L}\):
\begin{equation}
\begin{aligned}
c^{1:L}
&=
E_c(\phi_c(g_c^{1:L})),
\\
d^{1:L}
&=
E_d(\phi_d(g_d^{1:L})).
\end{aligned}
\label{eq:branch-evidence}
\end{equation}

To provide the routers with evidence quality cues, we estimate the consistency strength \(\kappa_c^\ell\) and conflict strength \(\kappa_d^\ell\) with learnable scorers \(\omega_c\) and \(\omega_d\):
\begin{equation}
\kappa_c^{\ell}
=
\sigma(\omega_c(c^{\ell})),
\quad
\kappa_d^{\ell}
=
\sigma(\omega_d(d^{\ell})).
\label{eq:evidence-strength}
\end{equation}
Together with the modality relevance and layer knowledge strength from the previous module, these scores form the reliability vector \(r^\ell\):
\begin{equation}
r^{\ell}
=
[
\mu_t^{\ell};
\mu_v^{\ell};
\nu_t^{\ell};
\nu_v^{\ell};
\kappa_c^{\ell};
\kappa_d^{\ell}
].
\label{eq:routing-reliability}
\end{equation}

For branch \(x\in\{c,d\}\), an independent router \(\psi_x\) computes the layer importance score \(\eta_x^\ell\) from the branch evidence \(x^\ell\), the decision query \(q\), and the reliability vector \(r^\ell\):
\begin{equation}
\eta_x^{\ell}
=
\psi_x([x^{\ell};q;r^{\ell}]).
\label{eq:route-logits}
\end{equation}
The router keeps the top \(K\) layers for each branch:
\begin{equation}
\mathcal{K}_x
=
\TopK(\{\eta_x^\ell\}_{\ell=1}^{L},K).
\label{eq:route-topk}
\end{equation}
The routing weights are obtained by normalizing the retained layer scores:
\begin{equation}
\alpha_x^\ell
=
\frac{
\exp(\eta_x^\ell)\mathbb{I}[\ell\in\mathcal{K}_x]
}{
\sum_{j\in\mathcal{K}_x}\exp(\eta_x^j)
}.
\label{eq:route-weights}
\end{equation}
The routed branch evidence \(\bar{x}\) is computed as
\begin{equation}
\bar{x}
=
\sum_{\ell=1}^{L}\alpha_x^{\ell}x^{\ell},
\quad
x\in\{c,d\}.
\label{eq:routed-evidence}
\end{equation}
Thus, \(\bar{c}\) and \(\bar{d}\) denote the routed consistency evidence and routed conflict evidence, respectively. The routed reliability summary \(\bar{r}\) averages the reliability vectors under the two routing distributions:
\begin{equation}
\bar{r}
=
\frac{1}{2}
\left(
\sum_{\ell=1}^{L}\alpha_c^{\ell}r^{\ell}
+
\sum_{\ell=1}^{L}\alpha_d^{\ell}r^{\ell}
\right).
\label{eq:routed-reliability}
\end{equation}

Finally, the classifier input \(z_{\mathrm{cls}}\) combines routed evidence, evidence interaction, evidence difference, the decision query, and the routed reliability summary:
\begin{equation}
z_{\mathrm{cls}}
=
[
\bar{c};
\bar{d};
\bar{c}\odot\bar{d};
|\bar{c}-\bar{d}|;
q;
\bar{r}
].
\label{eq:classifier-input}
\end{equation}
The fake news logit \(o\) and prediction probability \(\hat{y}\) are produced by
\begin{equation}
o=\mathrm{MLP}(z_{\mathrm{cls}}),
\quad
\hat{y}=\sigma(o).
\label{eq:prediction}
\end{equation}

\subsection{Training Objective}

We train the evidence network with the standard binary cross entropy loss \(\mathcal{L}_{\mathrm{cls}}\). To encourage the consistency and conflict streams to encode complementary evidence, we further use a disentanglement regularizer \(\mathcal{L}_{\mathrm{dis}}\):
\begin{equation}
\mathcal{L}_{\mathrm{dis}}
=
\frac{1}{B}
\sum_{b=1}^{B}
\frac{1}{L}
\sum_{\ell=1}^{L}
\cos(c_b^{\ell},d_b^{\ell})^2,
\label{eq:disentangle-loss}
\end{equation}
This term discourages the two streams from producing overly similar representations at each layer. With the disentanglement weight \(\lambda\), the final objective is
\begin{equation}
\mathcal{L}
=
\mathcal{L}_{\mathrm{cls}}
+
\lambda\mathcal{L}_{\mathrm{dis}}.
\label{eq:training-objective}
\end{equation}
During training, the LoRA tuned MLLM remains frozen, and only the three proposed modules and the classifier are optimized

\section{Experiments}

\subsection{Datasets and Metrics}

We evaluate MLCCD on three widely used multimodal fake news detection benchmarks: Weibo \cite{jin2017multimodal}, Pheme \cite{zubiaga2017exploiting}, and AMG \cite{guo2025each}. These datasets cover different languages, event sources, and data scales, enabling evaluation under diverse multimodal news verification scenarios. Table~\ref{tab:dataset} summarizes the dataset statistics used in our experiments. The dash in the Dev column indicates that Weibo is evaluated with its original train/test split and does not provide a separate development set. For all datasets, we report Accuracy, macro F1, real class F1, and fake class F1 as the main evaluation metrics. The detailed preprocessing procedures and split settings are provided in Appendix~\ref{app:dataset-details}.

\subsection{Baselines}

We compare MLCCD with three groups of baselines listed in Table~\ref{tab:main-results}. Among conventional multimodal fake news detectors, CFFN \citep{li2023cross} models consistency and inconsistency between words and image regions. MTS \citep{sun2025multimodal} captures multimodal interactions of different orders through Taylor series expansion, while MIMoE-FND \citep{liu2025modality} uses a hierarchical mixture of experts to adapt fusion to different modality interactions. DDU \citep{chen2026ddu} addresses epistemic and aleatoric uncertainty through retrieved context and uncertainty aware expert routing. We also include three general multimodal methods adapted to fake news detection: DMD \citep{li2023decoupled} decouples multimodal representations for cross modal knowledge distillation; DecAlign \citep{qian2026decalign} combines representation decoupling with hierarchical cross modal alignment; and MultiLoReFT \citep{tonekaboni2026multiloreft} learns shared and modality specific subspaces through low rank representation fine tuning. For MLLM baselines, FKA-Owl \citep{liu2024fka} augments large vision language models with forgery specific knowledge. We further construct direct MLLM baselines by LoRA fine tuning Gemma-4-E2B-it and Gemma-4-E4B-it \citep{gemma4modelcard}, as well as Qwen3-VL-2B-Instruct and Qwen3-VL-8B-Instruct \citep{bai2025qwen3}, under the same data splits.

\subsection{Implementation Details}

We implement MLCCD with Qwen3-VL-8B-Instruct as the backbone MLLM. The backbone is first adapted with LoRA; during evidence network training, the LoRA tuned backbone is frozen and only the proposed modules and the classifier are optimized. Each sample is formatted as one image and one text prompt. We train the evidence network with AdamW. For token selection, we set \(\rho_e=0.5\), keep at least one token for each non-empty top-ratio candidate set, fix \(\rho_q^v=0.3\), and set \(\rho_q^t\) to \(0.3\), \(0.025\), and \(0.1\) on Weibo, AMG, and Pheme, respectively. Detailed hyperparameters and input processing settings are provided in Appendix~\ref{app:implementation-details}.

\subsection{Overall Performance}

Table~\ref{tab:main-results} reports the overall results. MLCCD achieves the highest accuracy and macro F1 on all three datasets, outperforming the evaluated conventional detectors, general multimodal methods, and large model based baselines. Qwen3-VL-8B + LoRA is the strongest baseline in both metrics on each dataset. Compared with this baseline, MLCCD improves accuracy by \(0.90\), \(1.56\), and \(0.26\) percentage points on AMG, Pheme, and Weibo, respectively, with corresponding macro F1 gains of \(0.87\), \(2.27\), and \(0.26\) points. These results show additional gains over direct LoRA adaptation of the same backbone. The gains are modest on Weibo, where the baseline already achieves \(95.02\%\) in both metrics, and larger on AMG and Pheme.

The advantage also holds when compared with conventional fake news detectors and general multimodal methods adapted to this task. Among conventional detectors, DDU achieves the highest accuracy on AMG and Weibo, while MTS performs best on Pheme. MLCCD exceeds these results by \(4.98\), \(3.64\), and \(4.35\) accuracy points on AMG, Pheme, and Weibo, respectively. Among the general multimodal baselines, DMD achieves higher accuracy and macro F1 than DecAlign and MultiLoReFT on all three datasets, while MLCCD further improves on both metrics across these datasets.

Class wise results further show that the improvements are reflected in both labels, with MLCCD achieving the highest F1-R and F1-F on all three datasets. The largest gain over the strongest baseline in F1-F occurs on Pheme, where MLCCD reaches \(88.18\%\), compared with \(84.96\%\) for Gemma-4-E2B-it + LoRA, an improvement of \(3.22\) percentage points.

\begin{table}[t]
  \centering
  \small
  \caption{Dataset statistics.}
  \begin{tabular}{lrrrr}
    \toprule
    Dataset & Train & Dev & Test & Lang. \\
    \midrule
    Weibo & 7461 & -- & 1929 & Chinese \\
    Pheme & 1412 & 221 & 385 & English \\
    AMG & 3268 & 501 & 883 & English \\
    \bottomrule
  \end{tabular}
  \label{tab:dataset}
\end{table}

\begin{table}[!htbp]
  \centering
  \small
  \caption{Ablation results on Pheme. All values are percentages. Best results are in bold. F1-R and F1-F denote real class F1 and fake class F1.}
  \begin{tabular}{lcccc}
    \toprule
    Variant & Acc. & F1-Ma & F1-R & F1-F \\
    \midrule
    w/o TRTS & 92.21 & 90.45 & 94.55 & 86.36 \\
    w/o DLKE & 92.21 & 90.70 & 94.44 & 86.96 \\
    w/o ACCR & 92.47 & 90.70 & 94.76 & 86.64 \\
    Shared Router & 91.69 & 89.46 & 94.31 & 84.62 \\
    w/o Dis. & 92.73 & 91.09 & 94.91 & 87.27 \\
    MLCCD & \textbf{93.25} & \textbf{91.73} & \textbf{95.27} & \textbf{88.18} \\
    \bottomrule
  \end{tabular}

  \label{tab:ablation}
\end{table}

\begin{table*}[!t]
  \centering
  \small
  \caption{Overall results on AMG, PHEME, and Weibo with seed 42.
  All values are percentages. Acc. and F1-Ma denote accuracy and
  macro F1, respectively. F1-R and F1-F denote real class F1 and
  fake class F1, respectively.}
  \setlength{\tabcolsep}{2pt}
  \begin{tabular*}{\textwidth}{@{\extracolsep{\fill}}ll*{12}{c}@{}}
    \toprule
    \multirow{2}{*}{Category}
    & \multirow{2}{*}{Method}
    & \multicolumn{4}{c}{AMG}
    & \multicolumn{4}{c}{PHEME}
    & \multicolumn{4}{c}{Weibo} \\
    \cmidrule(lr){3-6}\cmidrule(lr){7-10}\cmidrule(lr){11-14}
    & & Acc. & F1-Ma & F1-R & F1-F
      & Acc. & F1-Ma & F1-R & F1-F
      & Acc. & F1-Ma & F1-R & F1-F \\
    \midrule

    \multirow{4}{*}{\shortstack[l]{General fake\\news detection\\methods}}
    & CFFN \citep{li2023cross}
    & 82.90 & 82.39 & 85.38 & 79.40
    & 82.86 & 79.22 & 87.91 & 70.54
    & 89.22 & 89.21 & 89.52 & 88.90 \\
    & MTS \citep{sun2025multimodal}
    & 85.05 & 84.51 & 87.40 & 81.62
    & 89.61 & 87.72 & 92.54 & 82.91
    & 89.58 & 89.54 & 90.15 & 88.94 \\
    & MIMoE-FND \citep{liu2025modality}
    & 84.03 & 83.64 & 86.16 & 81.12
    & 88.57 & 86.62 & 91.73 & 81.51
    & 89.89 & 89.88 & 90.19 & 89.58 \\
    & DDU \citep{chen2026ddu}
    & 85.39 & 84.91 & 87.61 & 82.21
    & 86.49 & 83.88 & 90.37 & 77.39
    & 90.93 & 90.93 & 91.02 & 90.83 \\
    \midrule

    \multirow{3}{*}{\shortstack[l]{General\\multimodal\\methods}}
    & DMD \citep{li2023decoupled}
    & 84.60 & 84.25 & 86.59 & 81.91
    & 88.57 & 85.24 & 92.25 & 78.22
    & 89.42 & 89.41 & 89.75 & 89.08 \\
    & DecAlign \citep{qian2026decalign}
    & 82.33 & 82.04 & 84.34 & 79.74
    & 86.23 & 83.61 & 90.17 & 77.06
    & 83.46 & 83.46 & 83.32 & 83.60 \\
    & MultiLoReFT \citep{tonekaboni2026multiloreft}
    & 79.50 & 79.21 & 81.66 & 76.77
    & 78.70 & 77.38 & 82.85 & 71.92
    & 84.71 & 84.67 & 85.43 & 83.91 \\
    \midrule

    \multirow{6}{*}{\shortstack[l]{MLLM\\methods}}
    & FKA-Owl \citep{liu2024fka}
    & 84.60 & 84.31 & 86.43 & 82.20
    & 85.97 & 83.50 & 89.89 & 77.12
    & 90.31 & 90.30 & 90.55 & 90.05 \\
    & Gemma-4-E2B-it + LoRA \citep{gemma4modelcard}
    & 84.94 & 84.62 & 86.82 & 82.43
    & 91.17 & 89.35 & 93.75 & 84.96
    & 91.81 & 91.81 & 91.96 & 91.66 \\
    & Gemma-4-E4B-it + LoRA \citep{gemma4modelcard}
    & 85.84 & 85.54 & 87.64 & 83.44
    & 91.17 & 89.24 & 93.80 & 84.68
    & 94.04 & 94.03 & 94.25 & 93.81 \\
    & Qwen3-VL-2B + LoRA \citep{bai2025qwen3}
    & 89.35 & 89.03 & 90.91 & 87.16
    & 90.13 & 88.34 & 92.91 & 83.76
    & 94.19 & 94.19 & 94.40 & 93.97 \\
    & Qwen3-VL-8B + LoRA \citep{bai2025qwen3}
    & 89.47 & 89.20 & 90.89 & 87.52
    & 91.69 & 89.46 & 94.31 & 84.62
    & 95.02 & 95.02 & 95.16 & 94.88 \\
    & Ours (MLCCD)
    & 90.37 & 90.07 & 91.80 & 88.34
    & 93.25 & 91.73 & 95.27 & 88.18
    & 95.28 & 95.28 & 95.34 & 95.22 \\
    \bottomrule
  \end{tabular*}

  \label{tab:main-results}
\end{table*}

\subsection{Ablation Study}

We conduct ablation studies on Pheme to examine the contribution of each component under the same training protocol. \textbf{w/o TRTS} removes Task Relevant Token Selection and uses all modality tokens as the candidate evidence pool. \textbf{w/o DLKE} removes Discriminative Layer Knowledge Extraction and does not select tokens according to layer transition scores. \textbf{w/o ACCR} removes Adaptive Consistency and Conflict Routing, and directly averages the layer aware textual and visual representations for prediction. \textbf{Shared Router} keeps the consistency and conflict streams but uses one shared router for both branches. \textbf{w/o Dis.} removes the disentanglement regularizer \(\mathcal{L}_{\mathrm{dis}}\).

Table~\ref{tab:ablation} shows that all ablated variants have lower accuracy and macro F1 than the full model. Removing TRTS or DLKE reduces macro F1 by \(1.28\) and \(1.03\) percentage points, respectively, suggesting that both task relevant token selection and layer transition based evidence extraction contribute to detection performance on Pheme. Removing ACCR also lowers macro F1 by \(1.03\) points, suggesting that the combined design of evidence decoupling and adaptive aggregation is more effective than directly averaging the layer representations. The shared router variant obtains the largest macro F1 drop of \(2.27\) points and reduces fake news F1 by \(3.56\) points, suggesting that consistency and conflict evidence benefit from separate layer routing in this setting. Removing \(\mathcal{L}_{\mathrm{dis}}\) leads to a smaller macro F1 drop of \(0.64\) points, suggesting that the regularizer also contributes to detection performance.

\subsection{Hyperparameter Sensitivity}

\begin{figure}[t]
  \centering
  \includegraphics[width=\columnwidth]{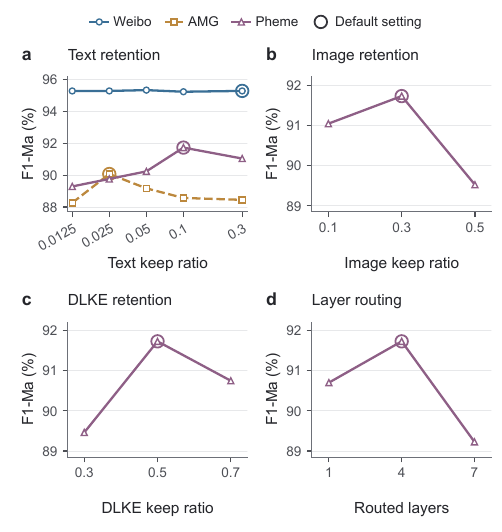}
  \caption{Hyperparameter sensitivity of MLCCD.
  (a) Text keep ratio $\rho_q^t$ on Weibo, AMG, and Pheme,
  shown on a logarithmic horizontal axis.
  (b) Image keep ratio $\rho_q^v$,
  (c) token keep ratio $\rho_e$ in DLKE, and
  (d) number of routed layers on Pheme.
  Outer circles indicate the default settings.
  Panels (b)--(d) share the same vertical scale.
  Each point represents one reported experimental result,
  and lines connect the evaluated settings.}
  \label{fig:hyperparameter-sensitivity}
\end{figure}

We examine the sensitivity of MLCCD to the text keep ratio \(\rho_q^t\) on all three datasets. On Pheme, we also evaluate the image keep ratio \(\rho_q^v\), the token keep ratio \(\rho_e\) in DLKE, and the number of routed layers. Figure~\ref{fig:hyperparameter-sensitivity} presents the macro F1 results, with outer circles indicating the default settings. Complete results for all four metrics are provided in Appendix~\ref{app:hyperparameter-sensitivity}.

\begin{table*}[t]
  \centering
  \small
  \caption{Results on the controlled counterfactual evaluation set. N denotes the number of samples, and Correct denotes the number assigned the intended label. Rates are percentages: counterfactual detection rates in the upper block and correct acceptance rates for real samples in the lower block. \(\Delta\) denotes the rate of MLCCD minus that of the LoRA baseline in percentage points.}
  \begin{tabular}{lrrrrrr}
    \toprule
    \multirow{2}{*}{Sample type}
      & \multirow{2}{*}{N}
      & \multicolumn{2}{c}{Qwen3-VL-8B + LoRA}
      & \multicolumn{2}{c}{MLCCD}
      & \multirow{2}{*}{\(\Delta\) (pp)} \\
    \cmidrule(lr){3-4}
    \cmidrule(lr){5-6}
      & & Correct & Rate (\%)
      & Correct & Rate (\%) & \\
    \midrule
    \multicolumn{7}{l}{Counterfactual samples: detection as fake} \\
    Object conflict
      & 100 & 23 & 23.00 & 50 & 50.00 & +27.00 \\
    Scene conflict
      & 100 & 23 & 23.00 & 52 & 52.00 & +29.00 \\
    Event conflict
      & 100 & 14 & 14.00 & 34 & 34.00 & +20.00 \\
    Claim conflict
      & 100 & 14 & 14.00 & 33 & 33.00 & +19.00 \\
    All counterfactuals
      & 400 & 74 & 18.50 & 169 & 42.25 & +23.75 \\
    \midrule
    \multicolumn{7}{l}{Real samples: correct acceptance as real} \\
    Original news
      & 100 & 97 & 97.00 & 96 & 96.00 & -1.00 \\
    Reference descriptions
      & 100 & 100 & 100.00 & 100 & 100.00 & 0.00 \\
    Meaning preserving local edits
      & 400 & 400 & 100.00 & 400 & 100.00 & 0.00 \\
    Semantic paraphrases
      & 100 & 100 & 100.00 & 100 & 100.00 & 0.00 \\
    \bottomrule
  \end{tabular}

  \label{tab:counterfactual-results}
\end{table*}

As shown in Figure~\ref{fig:hyperparameter-sensitivity}(a), the effect of the text keep ratio differs across datasets. On Weibo, macro F1 varies only from \(95.23\%\) to \(95.33\%\) over the tested ratios. The default ratio of \(0.3\) achieves \(95.28\%\), slightly below the highest value obtained at \(0.05\). In contrast, AMG and Pheme show greater variation. Among the tested values, AMG achieves its highest macro F1 of \(90.07\%\) at \(0.025\), while Pheme reaches \(91.73\%\) at \(0.1\). Both larger and smaller ratios yield lower macro F1 on these two datasets. These results show that the sensitivity to text token retention varies across datasets, with Weibo exhibiting relatively little variation over the tested range.

For the remaining parameters on Pheme, Figure~\ref{fig:hyperparameter-sensitivity}(b)--(d) shows that the default settings of \(\rho_q^v=0.3\), \(\rho_e=0.5\), and four routed layers achieve the highest macro F1 in their respective sweeps. Increasing or decreasing either token keep ratio from its default value lowers macro F1 within the tested ranges. Similarly, using one or seven routed layers yields macro F1 scores of \(90.70\%\) and \(89.23\%\), respectively, compared with \(91.73\%\) for four layers. These comparisons support the adopted settings on Pheme and show that retaining more tokens or aggregating more layers does not necessarily improve performance.

\subsection{Counterfactual Evaluation}

We conduct a controlled counterfactual evaluation to examine whether MLCCD can identify local conflicts within otherwise consistent multimodal content. Using images from 100 real news samples in the Pheme test set, we construct a reference description for each image and create four counterfactual variants involving objects, scenes, events, and claims. Each variant modifies a single contiguous text span while keeping the image and the remaining text unchanged, yielding 400 counterfactual samples. These edits are made to constructed reference descriptions rather than the original news text. For each group, we also retain the original news and include four meaning preserving local edits and one semantic paraphrase of the reference description. We use GPT-6.1 Sol (\texttt{gpt-6.1-sol}) through Codex as a tool
to assist in constructing reference descriptions and counterfactual
edits. The construction instructions and control generation
procedure are provided in
Appendix~\ref{app:counterfactual-construction}.

Table~\ref{tab:counterfactual-results} compares MLCCD with Qwen3-VL-8B + LoRA. For counterfactual samples, we report the detection rate, defined as the percentage classified as fake. For real samples, we report the correct acceptance rate, defined as the percentage classified as real. MLCCD detects 169 of the 400 counterfactual samples, compared with 74 for the LoRA baseline, increasing the overall detection rate from \(18.50\%\) to \(42.25\%\), a gain of \(23.75\) percentage points. Improvements are observed across all four conflict types. Both methods correctly classify all reference descriptions, meaning preserving local edits, and semantic paraphrases as real. On the 100 original news samples, MLCCD and the LoRA baseline correctly classify 96 and 97 samples, respectively.

These results indicate that MLCCD is more sensitive to the introduced local conflicts than direct LoRA fine tuning, while maintaining correct predictions on the constructed reference descriptions and meaning preserving controls. Across the four conflict types, MLCCD achieves higher detection rates for object and scene variants than for event and claim variants, indicating that these variants remain more difficult within this controlled evaluation.

\subsection{Analysis of Conflict Sparsity}

We further analyze the 400 counterfactual samples to examine detection performance across different degrees of conflict sparsity. For each sample, we define the edit ratio \(r_{\mathrm{edit}}\) as the number of tokens overlapping the changed text span divided by the total number of tokens in the counterfactual text body. The changed span is the smallest contiguous span containing the differences from the reference description. Token counts are obtained using the backbone tokenizer, excluding prompt, image, and padding tokens. A smaller ratio therefore indicates that the modification occupies a smaller portion of the text. We group the samples into three intervals and compute detection rates using the predictions from the counterfactual evaluation.

\begin{table}[t]
  \centering
  \small
  \caption{Counterfactual detection results grouped by edit ratio. N denotes the number of samples. Each method entry reports the number of detected counterfactuals, followed by the detection rate (\%) in parentheses. LoRA denotes Qwen3-VL-8B + LoRA.}
  \begin{tabular}{lccc}
    \toprule
    Edit ratio & N & LoRA & MLCCD \\
    \midrule
    \(r_{\mathrm{edit}}\leq5\%\)
      & 151 & 23 (15.23) & 54 (35.76) \\
    \(5\%<r_{\mathrm{edit}}\leq10\%\)
      & 153 & 29 (18.95) & 65 (42.48) \\
    \(10\%<r_{\mathrm{edit}}\leq20\%\)
      & 96 & 22 (22.92) & 50 (52.08) \\
    \midrule
    All samples
      & 400 & 74 (18.50) & 169 (42.25) \\
    \bottomrule
  \end{tabular}

  \label{tab:conflict-sparsity}
\end{table}

As shown in Table~\ref{tab:conflict-sparsity}, MLCCD achieves higher detection rates than the LoRA baseline in all three groups. When the edited span occupies at most \(5\%\) of the text body, MLCCD detects 54 of 151 counterfactual samples, compared with 23 for LoRA. The corresponding detection rates are \(35.76\%\) and \(15.23\%\), respectively, giving a difference of \(20.53\) percentage points. For the other two intervals, MLCCD achieves detection rates of \(42.48\%\) and \(52.08\%\), compared with \(18.95\%\) and \(22.92\%\) for LoRA. Both methods exhibit higher detection rates in groups with larger edit ratios.

\subsection{Analysis of Image Dependence}

To examine whether MLCCD's responses to local conflicts depend on visual content, we replace the original images while keeping the texts and model parameters unchanged. We consider two replacements: a uniform gray image with the same dimensions as the original image, and a donor image with the same number of visual tokens after preprocessing. We group samples by event using keyword rules applied to the source news text and select donor images from a different group. The gray image experiment covers all 100 groups, whereas suitable donor images are available for 64 groups, yielding 400 and 256 counterfactual and control pairs, respectively.

For each counterfactual text, we use its corresponding meaning preserving control and define the paired score difference as
\begin{equation}
    d(I) = s(I,T_{\mathrm{cf}}) - s(I,T_{\mathrm{ctrl}}),
\end{equation}
where \(s\) denotes the fake news logit produced by MLCCD, and \(T_{\mathrm{cf}}\) and \(T_{\mathrm{ctrl}}\) denote the counterfactual text and its control. A larger difference indicates a stronger relative response to the counterfactual text. We compute the mean difference under the original and replacement images using the same pairs within each experiment. This analysis measures changes in model responses, since the original compatibility labels do not necessarily apply after image replacement.

\begin{table}[t]
  \centering
  \small
  \caption{Effect of image replacement on MLCCD. N denotes the number of counterfactual and control pairs. Original and Replaced report the mean paired score difference under the respective image conditions, evaluated on the same pairs in each row.}
  \setlength{\tabcolsep}{4pt}
  \begin{tabular}{lrrr}
    \toprule
    Image replacement & N & Original & Replaced \\
    \midrule
    Uniform gray & 400 & 4.84 & 0.96 \\
    Donor image & 256 & 4.70 & 1.02 \\
    \bottomrule
  \end{tabular}

  \label{tab:image-dependence}
\end{table}

As shown in Table~\ref{tab:image-dependence}, replacing the original images with uniform gray images reduces the mean paired score difference from \(4.84\) to \(0.96\). On the subset with matched donor images, the difference decreases from \(4.70\) to \(1.02\). The reduction under both replacements indicates that visual content contributes to the separation between counterfactual texts and meaning preserving controls in this evaluation.

\subsection{Efficiency Analysis}
Fake news can spread rapidly through online platforms \cite{khattar2019mvae}, making timely detection important for practical deployment. As new content continually arrives, deployed detectors must repeatedly perform inference, so its efficiency directly affects how quickly incoming news can be assessed. We therefore compare the training and inference times of MLCCD with Qwen3-VL-8B + LoRA on Pheme using execution logs. Both training stages use four NVIDIA A100 80GB GPUs with a global batch size of \(24\), while inference for both methods is performed on the same single GPU with a batch size of \(24\). Total training time covers five epochs of LoRA adaptation and, for MLCCD, five additional epochs of evidence network training, excluding model initialization, validation, and checkpoint saving.

\begin{table}[t]
  \centering
  \small
  \caption{Runtime comparison on Pheme. MLCCD training time includes both LoRA adaptation and evidence network training. Inference time covers all 385 test samples. Timings are derived from execution logs and include data handling and logging overhead.}
  \setlength{\tabcolsep}{4pt}
  \begin{tabular}{lccc}
    \toprule
    Method
    & \shortstack{Total training\\time (s)}
    & \shortstack{Training\\overhead}
    & \shortstack{Inference\\time (s)} \\
    \midrule
    Qwen3-VL-8B + LoRA
    & 845
    & --
    & \(\approx 45\) \\
    MLCCD
    & 1102
    & \shortstack{+257 s\\(+30.4\%)}
    & \(\approx 48\) \\
    \bottomrule
  \end{tabular}

  \label{tab:efficiency}
\end{table}

As shown in Table~\ref{tab:efficiency}, MLCCD requires \(1{,}102\) seconds of training, including \(845\) seconds for LoRA adaptation and \(257\) seconds for evidence network training. This corresponds to a \(30.4\%\) increase over direct LoRA fine tuning. During inference, however, MLCCD processes all \(385\) test samples in approximately \(48\) seconds, compared with \(45\) seconds for the baseline, adding only about \(3\) seconds across the entire test set. With this small increase in inference time, MLCCD improves accuracy by \(1.56\) percentage points and macro F1 by \(2.27\) percentage points on Pheme. These results show that MLCCD offers a favorable balance between predictive performance and processing efficiency for timely news verification on the Web.

\section{Conclusion}

We proposed MLCCD, a framework that draws on knowledge encoded across different MLLM layers for multimodal fake news detection. MLCCD first selects task-relevant visual and textual tokens and then retains representations that change substantially between adjacent layers to capture complementary information. It further separates consistency and conflict evidence and aggregates them through two adaptive routers to address sparse conflicts and varying evidence needs. Experiments on three benchmarks demonstrate improvements over the evaluated baselines. Controlled counterfactual evaluation further shows that MLCCD detects a larger proportion of local conflicts than direct LoRA fine-tuning while maintaining low false-alarm rates on the evaluated controls. Timing results on PHEME indicate limited additional inference cost, supporting the potential of layer-specific evidence modeling for assessing the credibility of multimodal Web content.

\bibliographystyle{ACM-Reference-Format}
\bibliography{sample-base}

\appendix
\section{Dataset Details}
\label{app:dataset-details}

We evaluate MLCCD on three multimodal fake news benchmarks: Weibo \cite{jin2017multimodal}, Pheme \cite{zubiaga2017exploiting}, and AMG \cite{guo2025each}. We follow the standard split and preprocessing protocol of each dataset where available. For Weibo, the original train/test split is kept, and when a post contains multiple images, the image with the highest resolution is used as the visual input. Following common preprocessing practice \cite{chen2022cross,lao2024frequency}, we remove samples with empty text, including 50 training samples and 66 test samples, as well as one test sample without a valid image. For Pheme, we adopt the same train/validation/test split as \cite{zheng2022mfan}. For AMG, we remove samples whose media files are not images, corrupted images, and images with abnormal resolution such as \(1\times1\). Since AMG contains substantially longer texts, we truncate each AMG text to the first 1024 words during preprocessing. Table~\ref{tab:dataset} reports the resulting statistics. For evaluation, we report Accuracy, macro F1, real class F1, and fake class F1.

\section{Implementation Details}
\label{app:implementation-details}

For LoRA adaptation, we use rank \(256\), LoRA alpha \(512\), dropout \(0.05\), and apply LoRA to the query, key, value, output, gate, up, and down projection layers. For Qwen based models, the processor uses left padding and constrains the visual token budget to \(4\)--\(256\) tokens by setting \(\texttt{min\_pixels}=4\times32^2\) and \(\texttt{max\_pixels}=256\times32^2\). For Gemma based baselines, we set the maximum visual soft token budget to \(280\), the supported budget closest to the Qwen limit of \(256\).

The evidence network follows a compact configuration. We set the hidden dimension to \(1024\) and the classifier fusion dimension to \(512\). Each lightweight Transformer encoder contains one block with \(8\) attention heads and dropout \(0.1\). For token selection based on changes between adjacent layers, we set \(\rho_e=0.5\). Each top ratio operation keeps at least one token when the candidate set is non-empty. The image keep ratio is fixed as \(\rho_q^v=0.3\) across datasets. We set the text keep ratio \(\rho_q^t\) to \(0.3\) on Weibo, \(0.025\) on AMG, and \(0.1\) on Pheme. During routing, each evidence stream keeps the top \(4\) layers. We train the evidence network for \(5\) epochs with AdamW, using learning rate \(5\times10^{-5}\), weight decay \(0.01\), global batch size \(24\), gradient clipping \(1.0\), and DeepSpeed ZeRO stage 2 with bfloat16. We use cosine learning rate decay with a warmup ratio of \(0.1\). The weight of the disentanglement loss is set to \(0.01\).

\section{Hyperparameter Sensitivity}
\label{app:hyperparameter-sensitivity}

Tables~\ref{tab:text-keep-sensitivity} and~\ref{tab:pheme-hyperparameter-sensitivity} provide the complete results for the hyperparameter sweeps in Figure~\ref{fig:hyperparameter-sensitivity}. We report accuracy, macro F1, real class F1, and fake class F1 for every evaluated setting.

\begin{table*}[!t]
  \centering
  \small
  \caption{Sensitivity to the text keep ratio \(\rho_q^t\). All performance scores are percentages. The default ratios are \(0.3\), \(0.025\), and \(0.1\) for Weibo, AMG, and Pheme, respectively. F1-R and F1-F denote real class F1 and fake class F1.}
  \begin{tabular}{c*{12}{c}}
    \toprule
    \multirow{2}{*}{\(\rho_q^t\)}
    & \multicolumn{4}{c}{Weibo}
    & \multicolumn{4}{c}{AMG}
    & \multicolumn{4}{c}{Pheme} \\
    \cmidrule(lr){2-5}
    \cmidrule(lr){6-9}
    \cmidrule(lr){10-13}
    & Acc. & F1-Ma & F1-R & F1-F
    & Acc. & F1-Ma & F1-R & F1-F
    & Acc. & F1-Ma & F1-R & F1-F \\
    \midrule
    0.0125
    & 95.28 & 95.28 & 95.39 & 95.17
    & 88.45 & 88.26 & 89.74 & 86.79
    & 91.43 & 89.29 & 94.08 & 84.51 \\
    0.025
    & 95.28 & 95.28 & 95.36 & 95.20
    & 90.37 & 90.07 & 91.80 & 88.34
    & 91.69 & 89.76 & 94.20 & 85.32 \\
    0.05
    & 95.33 & 95.33 & 95.45 & 95.22
    & 89.47 & 89.17 & 90.96 & 87.38
    & 92.21 & 90.24 & 94.62 & 85.85 \\
    0.1
    & 95.23 & 95.23 & 95.29 & 95.17
    & 88.79 & 88.57 & 90.15 & 86.99
    & 93.25 & 91.73 & 95.27 & 88.18 \\
    0.3
    & 95.28 & 95.28 & 95.34 & 95.22
    & 88.67 & 88.45 & 90.06 & 86.84
    & 92.73 & 91.04 & 94.93 & 87.16 \\
    \bottomrule
  \end{tabular}

  \label{tab:text-keep-sensitivity}
\end{table*}

\begin{table}[!htbp]
  \centering
  \small
  \caption{Sensitivity to the image keep ratio \(\rho_q^v\), the token keep ratio \(\rho_e\) in DLKE, and the number of routed layers on Pheme. All performance scores are percentages. \(\dagger\) marks the default setting in each group. F1-R and F1-F denote real class F1 and fake class F1.}
  \setlength{\tabcolsep}{4pt}
  \begin{tabular}{lccccc}
    \toprule
    Parameter & Value & Acc. & F1-Ma & F1-R & F1-F \\
    \midrule
    \multirow{3}{*}{\(\rho_q^v\)}
    & 0.1 & 92.73 & 91.04 & 94.93 & 87.16 \\
    & \(0.3^\dagger\) & 93.25 & 91.73 & 95.27 & 88.18 \\
    & 0.5 & 91.69 & 89.52 & 94.29 & 84.76 \\
    \midrule
    \multirow{3}{*}{\(\rho_e\)}
    & 0.3 & 91.69 & 89.46 & 94.31 & 84.62 \\
    & \(0.5^\dagger\) & 93.25 & 91.73 & 95.27 & 88.18 \\
    & 0.7 & 92.47 & 90.75 & 94.74 & 86.76 \\
    \midrule
    \multirow{3}{*}{Routed layers}
    & 1 & 92.21 & 90.70 & 94.44 & 86.96 \\
    & \(4^\dagger\) & 93.25 & 91.73 & 95.27 & 88.18 \\
    & 7 & 91.43 & 89.23 & 94.10 & 84.36 \\
    \bottomrule
  \end{tabular}

  \label{tab:pheme-hyperparameter-sensitivity}
\end{table}

\section{Counterfactual Sample Construction}
\label{app:counterfactual-construction}

We use GPT-6.1 Sol (\texttt{gpt-6.1-sol}) through Codex to assist
in constructing reference descriptions and counterfactual edits
from Pheme images. Construction and visual review are performed
through an interactive workflow. The instructions and procedures for constructing the counterfactual samples are summarized below.

\begin{quote}
\small
You are assisting in constructing a controlled evaluation set for
multimodal fake news detection. Given a news image and its source
text, produce four reference clauses supported by the visible image
and a contradictory text replacement for each clause.

\begin{enumerate}
    \item Cover four edit types:
    object, concerning a visible object or entity;
    scene, concerning the visible environment;
    event, concerning a visible action, posture, state, or arrangement;
    and claim, concerning another directly checkable assertion,
    such as an attribute or clearly readable text.

    \item Base every reference clause on visible evidence.
    Use the source news only as context.
    Do not introduce identities, intentions, dates, locations,
    or historical outcomes that cannot be verified from the image.

    \item Write each reference clause using the word ``shown''
    to indicate that it describes the supplied image.
    Make the four clauses compatible with one another,
    and avoid repeating the same target fact across clauses.

    \item For each clause, identify one short, contiguous text span
    and propose a replacement that directly contradicts the image.
    The replacement must preserve grammatical coherence.
    All other text and the image must remain unchanged.

    \item Avoid edits that create an obvious contradiction between
    the textual clauses themselves.
    The intended inconsistency should depend on comparing the
    edited description with the image.

    \item For each edit, record its type, reference clause,
    original span, replacement span, and supporting visual evidence.
    If four unambiguous contrasts cannot be supported,
    reject the candidate image rather than inventing a conflict.
\end{enumerate}

Return structured records with the fields
\texttt{type},
\texttt{reference\_clause},
\texttt{changed\_from},
\texttt{changed\_to}, and
\texttt{image\_evidence}.
\end{quote}

Across the 100 groups, we balance the order of the four reference
clauses. The clause corresponding to each conflict type appears
in the first, second, third, and fourth sentence positions in
25 groups each. The source image remains unchanged.

The meaning preserving controls are generated programmatically.
Each local control replaces ``shown'' with ``pictured'' in one
reference clause. The full description control rewrites all four
clauses using the form ``In this image, ...'', while retaining
their factual content.

\section{Detailed Classification Results}
\label{app:detailed-results}

Tables~\ref{tab:detailed-amg}, \ref{tab:detailed-pheme}, and~\ref{tab:detailed-weibo} report accuracy and the precision, recall, and F1 score for each class on AMG, Pheme, and Weibo, respectively. All values are percentages. Prec. and Rec. denote precision and recall, respectively.

\begin{table}[t]
  \centering
  \small
  \caption{Detailed classification results on AMG.}
  \setlength{\tabcolsep}{1.5pt}
  \begin{tabular*}{\columnwidth}{@{\extracolsep{\fill}}lrrrrrrr@{}}
    \toprule
    \multirow{2}{*}{Method}
    & \multirow{2}{*}{Acc.}
    & \multicolumn{3}{c}{Fake News}
    & \multicolumn{3}{c}{Real News} \\
    \cmidrule(lr){3-5}\cmidrule(lr){6-8}
    & & Prec. & Rec. & F1 & Prec. & Rec. & F1 \\
    \midrule
    CFFN & 82.90 & 74.42 & 85.09 & 79.40 & 89.63 & 81.52 & 85.38 \\
    MTS & 85.05 & 77.93 & 85.67 & 81.62 & 90.34 & 84.66 & 87.40 \\
    MIMoE-FND & 84.03 & 74.81 & 88.60 & 81.12 & 91.84 & 81.15 & 86.16 \\
    DDU & 85.39 & 77.81 & 87.13 & 82.21 & 91.20 & 84.29 & 87.61 \\
    DMD & 84.60 & 75.12 & 90.06 & 81.91 & 92.81 & 81.15 & 86.59 \\
    DecAlign & 82.33 & 71.73 & 89.77 & 79.74 & 92.31 & 77.63 & 84.34 \\
    MultiLoReFT & 79.50 & 68.42 & 87.43 & 76.77 & 90.36 & 74.49 & 81.66 \\
    FKA-Owl & 84.60 & 74.41 & 91.81 & 82.20 & 93.93 & 80.04 & 86.43 \\
    Gemma-4-E2B-it + LoRA & 84.94 & 75.18 & 91.23 & 82.43 & 93.59 & 80.96 & 86.82 \\
    Gemma-4-E4B-it + LoRA & 85.84 & 76.27 & 92.11 & 83.44 & 94.26 & 81.89 & 87.64 \\
    Qwen3-VL-2B + LoRA & 89.35 & 81.79 & 93.27 & 87.16 & 95.33 & 86.88 & 90.91 \\
    Qwen3-VL-8B + LoRA & 89.47 & 80.89 & 95.32 & 87.52 & 96.67 & 85.77 & 90.89 \\
    MLCCD & 90.37 & 83.20 & 94.15 & 88.34 & 95.97 & 87.99 & 91.80 \\
    \bottomrule
  \end{tabular*}

  \label{tab:detailed-amg}
\end{table}

\begin{table}[t]
  \centering
  \small
  \caption{Detailed classification results on Pheme.}
  \setlength{\tabcolsep}{1.5pt}
  \begin{tabular*}{\columnwidth}{@{\extracolsep{\fill}}lrrrrrrr@{}}
    \toprule
    \multirow{2}{*}{Method}
    & \multirow{2}{*}{Acc.}
    & \multicolumn{3}{c}{Fake News}
    & \multicolumn{3}{c}{Real News} \\
    \cmidrule(lr){3-5}\cmidrule(lr){6-8}
    & & Prec. & Rec. & F1 & Prec. & Rec. & F1 \\
    \midrule
    CFFN & 82.86 & 71.17 & 69.91 & 70.54 & 87.59 & 88.24 & 87.91 \\
    MTS & 89.61 & 80.17 & 85.84 & 82.91 & 93.94 & 91.18 & 92.54 \\
    MIMoE-FND & 88.57 & 77.60 & 85.84 & 81.51 & 93.85 & 89.71 & 91.73 \\
    DDU & 86.49 & 76.07 & 78.76 & 77.39 & 91.04 & 89.71 & 90.37 \\
    DMD & 88.57 & 88.76 & 69.91 & 78.22 & 88.51 & 96.32 & 92.25 \\
    DecAlign & 86.23 & 75.42 & 78.76 & 77.06 & 91.01 & 89.34 & 90.17 \\
    MultiLoReFT & 78.70 & 58.66 & 92.92 & 71.92 & 96.12 & 72.79 & 82.85 \\
    FKA-Owl & 85.97 & 73.98 & 80.53 & 77.12 & 91.60 & 88.24 & 89.89 \\
    Gemma-4-E2B-it + LoRA & 91.17 & 84.96 & 84.96 & 84.96 & 93.75 & 93.75 & 93.75 \\
    Gemma-4-E4B-it + LoRA & 91.17 & 86.24 & 83.19 & 84.68 & 93.12 & 94.49 & 93.80 \\
    Qwen3-VL-2B + LoRA & 90.13 & 80.99 & 86.73 & 83.76 & 94.32 & 91.54 & 92.91 \\
    Qwen3-VL-8B + LoRA & 91.69 & 92.63 & 77.88 & 84.62 & 91.38 & 97.43 & 94.31 \\
    MLCCD & 93.25 & 90.65 & 85.84 & 88.18 & 94.24 & 96.32 & 95.27 \\
    \bottomrule
  \end{tabular*}

  \label{tab:detailed-pheme}
\end{table}

\begin{table}[t]
  \centering
  \small
  \caption{Detailed classification results on Weibo.}
  \setlength{\tabcolsep}{1.5pt}
  \begin{tabular*}{\columnwidth}{@{\extracolsep{\fill}}lrrrrrrr@{}}
    \toprule
    \multirow{2}{*}{Method}
    & \multirow{2}{*}{Acc.}
    & \multicolumn{3}{c}{Fake News}
    & \multicolumn{3}{c}{Real News} \\
    \cmidrule(lr){3-5}\cmidrule(lr){6-8}
    & & Prec. & Rec. & F1 & Prec. & Rec. & F1 \\
    \midrule
    CFFN & 89.22 & 88.52 & 89.28 & 88.90 & 89.88 & 89.16 & 89.52 \\
    MTS & 89.58 & 91.40 & 86.60 & 88.94 & 88.04 & 92.37 & 90.15 \\
    MIMoE-FND & 89.89 & 89.34 & 89.82 & 89.58 & 90.41 & 89.96 & 90.19 \\
    DDU & 90.93 & 88.83 & 92.93 & 90.83 & 93.07 & 89.06 & 91.02 \\
    DMD & 89.42 & 88.98 & 89.17 & 89.08 & 89.84 & 89.66 & 89.75 \\
    DecAlign & 83.46 & 80.34 & 87.14 & 83.60 & 86.91 & 80.02 & 83.32 \\
    MultiLoReFT & 84.71 & 85.44 & 82.42 & 83.91 & 84.06 & 86.85 & 85.43 \\
    FKA-Owl & 90.31 & 89.43 & 90.68 & 90.05 & 91.15 & 89.96 & 90.55 \\
    Gemma-4-E2B-it + LoRA & 91.81 & 90.32 & 93.03 & 91.66 & 93.29 & 90.66 & 91.96 \\
    Gemma-4-E4B-it + LoRA & 94.04 & 94.26 & 93.35 & 93.81 & 93.83 & 94.68 & 94.25 \\
    Qwen3-VL-2B + LoRA & 94.19 & 94.38 & 93.57 & 93.97 & 94.02 & 94.78 & 94.40 \\
    Qwen3-VL-8B + LoRA & 95.02 & 94.38 & 95.39 & 94.88 & 95.64 & 94.68 & 95.16 \\
    MLCCD & 95.28 & 93.31 & 97.21 & 95.22 & 97.28 & 93.47 & 95.34 \\
    \bottomrule
  \end{tabular*}

  \label{tab:detailed-weibo}
\end{table}

\clearpage

\end{document}